\RequirePackage{fix-cm}
\documentclass[a4paper,11pt]{article}

\usepackage{authblk}
\usepackage[utf8]{inputenc}
\usepackage[T1]{fontenc}
\usepackage{pslatex}
\usepackage[english]{babel}
\usepackage{fancyhdr}
\usepackage{hyperref}
\usepackage{amsmath,amssymb,amsfonts,amsthm}
\usepackage{graphicx}
\usepackage{mathtools}
\usepackage{algorithm}
\usepackage{algorithmic}
\usepackage{microtype}
\usepackage{rotating}
\usepackage{todonotes}
\usepackage{tikz}
\usepackage{tikz-qtree}
\usepackage{xparse}
\usepackage{appendix}
\usepackage{url}
\usepackage{tabularx}
\usepackage{paralist}
\usepackage{booktabs}
\usepackage{multirow}
\usepackage{subcaption}
\usepackage{cancel}
\usetikzlibrary{calc,shapes.callouts,shapes.arrows}
\usetikzlibrary{automata}
\usetikzlibrary{positioning}
\usetikzlibrary{decorations.pathreplacing}
\usetikzlibrary{decorations.pathmorphing}
\usetikzlibrary{arrows}
\usetikzlibrary{shapes}

\newcommand{\N}{\mathbb{N}}
\newcommand{\Z}{\mathbb{Z}}
\newcommand{\F}{\mathbb{F}}

\providecommand{\keywords}[1]{\textbf{\textit{Keywords }} #1}

\begin{document}

\title{Degree is Important: On Evolving Homogeneous Boolean Functions}

\author[1]{Claude Carlet}
\author[2]{Marko \DH urasevic}
\author[2]{Domagoj Jakobovic}
\author[3]{Luca Mariot}
\author[4]{Stjepan Picek}

\affil[1 ]{{\normalsize University of Bergen, Bergen, Norway}

    {\small \texttt{claude.carlet@gmail.com}}}

\affil[2 ]{{\normalsize Faculty of Electrical Engineering and Computing, University of Zagreb, Unska 3, Zagreb, Croatia} \\

{\small \texttt{marko.durasevic@fer.hr, domagoj.jakobovic@fer.hr}}}

\affil[3 ]{{\normalsize Semantics, Cybersecurity and Services Group, University of Twente, 7522 NB Enschede, The Netherlands}

{\small \texttt{l.mariot@utwente.nl}}}

\affil[4 ]{{\normalsize Digital Security Group, Radboud University, Postbus 9010, 6500 GL Nijmegen, The Netherlands}
	
	{\small \texttt{stjepan.picek@ru.nl}}}
	
\maketitle

\begin{abstract}
Boolean functions with good cryptographic properties like high nonlinearity and algebraic degree play an important in the security of stream and block ciphers. Such functions may be designed, for instance, by algebraic constructions or metaheuristics. This paper investigates the use of Evolutionary Algorithms (EAs) to design homogeneous bent Boolean functions, i.e., functions that are maximally nonlinear and whose algebraic normal form contains only monomials of the same degree. In our work, we evaluate three genotype encodings and four fitness functions. Our results show that while EAs manage to find quadratic homogeneous bent functions (with the best method being a GA leveraging a restricted encoding), none of the approaches result in cubic homogeneous bent functions.
\end{abstract}


\section{Introduction}
\label{sec:introduction}

Boolean functions satisfying specific properties have been an active research topic in symmetric cryptography for several decades already~\cite{carlet_2021}. Common requirements for the design of stream and block ciphers are that the underlying Boolean functions are balanced, highly nonlinear, have a specific algebraic degree, or are easy to implement, to list only a few relevant cases. There are several ways to obtain functions with good cryptographic properties. The first option is to use algebraic constructions, where one finds a way to construct an infinite family of Boolean functions having the desired properties. However, finding appropriate constructions can be very difficult. Moreover, there are classes of functions for which we do not know any algebraic construction, and we are forced to work with empirical approaches.

Common empirical approaches for this problem include random search, specific heuristics, and metaheuristics.
Random search is easy to implement, but it rarely gives satisfactory results except for very simple problems. Specific heuristics often perform very well, sometimes even giving the best-known results~\cite{4167738}. Depending on the problem, it can still perform unsatisfactorily, or it can be difficult to design a good algorithm.
Finally, metaheuristics offer a good trade-off between the aforesaid aspects as it often performs well (even optimal) and commonly does not require specialized knowledge about the problem~\cite{Djurasevic2023}.

This paper deals with a difficult problem where there is only a handful of known solutions. More precisely, we explore whether we can evolve homogeneous Boolean functions, which have the same algebraic degree for all monomials in their algebraic normal form representation. Such functions are interesting to consider as they may allow for a more efficient evaluation.

Up to now, there has been very little research on constructing homogeneous Boolean functions. The first work goes back to 2001, where the authors used combinatorial methods to find bent and balanced homogeneous functions in dimension 6~\cite{10.1007/3-540-48970-3_3}. 
We note that the characterization of quadratic homogeneous bent functions is well-known~\cite{MacWilliams-Sloane}. Interestingly, there is only one primary construction of cubic homogeneous bent functions~\cite{Seberry2000ConstructionOC}.
To our knowledge, no works up to now considered a metaheuristic approach for evolving homogeneous Boolean functions. On the other hand, evolving bent Boolean functions is a rather common research direction where multiple works and approaches were successful, see, e.g.,~\cite{FullerDM03, 10.1007/978-3-319-10762-2_41, 10.1145/2908812.2908915, manzoni20, mariot22}. While being less explored, evolving Boolean functions with a good algebraic degree (among other properties) is also done with good results~\cite{00190, cryptoeprint:2013:011, 10.1007/978-3-319-10762-2_81}. 

In this paper, we investigate two Boolean function representations (truth table and algebraic normal form) and three solution encodings (bitstring, bitstring restricted, and tree-based). First, we experiment with Boolean function sizes from dimension 6 to dimension 14 and evolve homogeneous Boolean functions (with no further constraints). Then, we explore evolving bent (maximally nonlinear) homogeneous Boolean functions.
Our results indicate that while evolving homogeneous Boolean functions is relatively easy, genetic programming finds such solutions for all tested Boolean sizes. On the other hand, when evolving homogeneous bent Boolean functions, the problem is already difficult even for small Boolean function sizes, and we manage to find only quadratic homogeneous bent functions where a genetic algorithm with restricted encoding works the best.

\section{Preliminaries}
\label{sec:background}
We denote by $\F_2=\{0,1\}$ the finite field with two elements, equipped with XOR and logical AND, respectively, as the sum and multiplication operations. The $n$-dimensional vector space over $\F_2$ is denoted as $\F_2^n$, consisting of all $2^n$ binary vectors of length $n$. Given $a, b \in \F_2^n$, their inner product equals $a\cdot b = \bigoplus_{i=1}^{n} a_{i}b_{i}$ in $\mathbb F_{2}^n$. A Boolean function of $n$ variables is a mapping $f: \F_2^n \to \F_2$. Interested readers can find detailed information about Boolean functions in~\cite{MacWilliams-Sloane,carlet_2021}.

\subsection{Boolean Function Representations}

\subsubsection{Truth Table Representation}
The most basic means to uniquely represent a Boolean function $f: \F_2^n \to \F_2$ is by using its truth table. The truth table of a Boolean function $f$ is the list of pairs $(x, f(x))$ of input vectors $x \in \F_2^n$ and function outputs $f(x) \in \F_2$. Once a total order has been fixed on the input vectors of $\F_2^n$ (most commonly, the lexicographic order), the truth table can be identified only by the $2^n$-bit function output vector.

\subsubsection{Walsh-Hadamard Transform}
The Walsh-Hadamard transform $W_{f}: \F_2^n \to \Z$ is another commonly used unique representation of a Boolean function $f: \F_2^n \to \F_2$, which allows to characterize several interesting cryptographic properties. Formally, the Walsh-Hadamard transform measures the correlation between $f$ and the linear functions $a\cdot x$, for all $a \in \N$, as follows:
\begin{equation}
W_{f} (a) = \sum\limits_{x \in \mathbb{F}_{2}^{n}} (-1)^{f(x) \oplus a\cdot x},
\end{equation}
with the sum calculated in ${\mathbb Z}$. The Walsh-Hadamard transform is an involution up to a normalization by a constant. Therefore, one can retrieve the truth table representation of $f$ from the spectrum of its Walsh-Hadamard coefficients $W_f(a)$.

\subsubsection{Algebraic Normal Form}
A third unique representation of a Boolean function $f: \F_2^n \to \F_2$ is as a multivariate polynomial in the quotient ring
$\mathbb{F}_{2}\left[x_{1},..., x_{n}\right]/(x_{1}^{2} \oplus x_{1},..., x_{n}^{2} \oplus x_{n})$. This polynomial is the Algebraic Normal Form (ANF) of $f$, and it is defined as:
\begin{equation}
f(x) = \bigoplus_{\substack{a \in \mathbb{F}_{2}^{n}}} h(a)\cdot x^{a},
\end{equation}  
\noindent
where $h(a)$ is given by the binary M\"{o}bius transform:
\begin{equation}
h(a)= \bigoplus_{\substack{x \preceq a}}  f(x), \text{ for any } a \in \mathbb{F}_2^n,
\end{equation}
with $\preceq$ denoting the covering relation between vectors of $\F_2^n$, i.e. $a$ covers $x$ means that $x_i \leq a_i, \forall i \in \left\lbrace 0, \ldots, n-1 \right\rbrace$.
Similarly to the Walsh-Hadamard transform, the M\"{o}bius transform is also an involution, so one can use it to switch between the truth table and the algebraic normal form representations of a function.

\subsection{Properties and Bounds}
\label{sec:boolean_properties}

\paragraph{Nonlinearity}
The minimum Hamming distance between a Boolean function $f$ and all affine functions is the nonlinearity of $f$.
The nonlinearity $nl_{f}$ of a Boolean function $f$ is calculated from the Walsh-Hadamard spectrum~\cite{carlet_2021}:
\begin{equation}
\label{eq:nonlinearity}
nl_{f} = 2^{n - 1} - \frac{1}{2}\max_{a \in \mathbb{F}_{2}^{n}} \left \{ |W_{f}(a)| \right \}.
\end{equation}

For every $n$-variable Boolean function, $f$ satisfies the so-called covering radius bound:
\begin{equation}
\label{eq_boolean_covering}
    nl_{f} \leq 2^{n-1}-2^{\frac n 2 - 1}.
\end{equation}
Notice that Eq.~\eqref{eq_boolean_covering} cannot be tight when $n$ is odd. 

\paragraph{Bentness}

The functions whose nonlinearity equals the maximal value from Eq.~\eqref{eq_boolean_covering} are called bent. Bent functions exist for $n$ even only.

\paragraph{Algebraic Degree}
The algebraic degree $deg_f$ of a Boolean function \textit{f} is defined as the number of variables in the largest product term of the function's ANF having a non-zero coefficient, see~\cite{MacWilliams-Sloane}:
\begin{equation}
deg_f = \max \{w_H(a) : \ a \in \F_2^n, \ h(a) = 1 \}.
\end{equation} 

The algebraic degree of a bent function cannot exceed $n/2$.
A Boolean function is affine if and only if it has an algebraic degree at most 1. We will call quadratic functions the Boolean functions of algebraic degree at most 2 and cubic functions those of algebraic degree at most 3. This means that an affine function is a particular quadratic function (in the same sense that a constant function is a particular affine function).

\paragraph{Homogeneity}
A Boolean function is called homogeneous if all the monomials in its algebraic normal form have the same algebraic degree. 
The only known homogeneous bent functions are quadratic and cubic ones. Furthermore, it is not known whether homogeneous bent functions of higher degrees exist~\cite{Polujan2020}. 

\section{Methodology}
\label{sec:methodology}

\subsection{Encoding}

\paragraph{Bitstring Encoding.}
The most common option for encoding a Boolean function is the bitstring (BS) encoding~\cite{Djurasevic2023}, representing the truth table of a Boolean function. 
For a Boolean function with $n$ inputs, the truth table is encoded as a bitstring of length $2^n$.
The bitstring represents the Boolean function upon which the algorithm operates directly. Therefore, the algorithm, in this case, explores the full space of $n$-variable Boolean functions, which has size $2^{2^n}$. In each evaluation, the truth table is transformed into the Walsh-Hadamard spectrum or the ANF form, after which the nonlinearity and homogeneity properties are evaluated.

\paragraph{Restricted Encoding.}
The bitstring encoding introduced above does not guarantee that the resulting function is always homogeneous when converted in its ANF representation. Hence, we also considered a second encoding based on the ANF, where an individual is represented as a bitstring of length $\binom{n}{d}$, with $d$ being the target degree. Each component in this bitstring specifies whether the corresponding $d$-degree monomial occurs in the ANF of the function (1) or not (0). Standard genetic operators can be applied to this bitstring, and the individual is then mapped to the full ANF representation by setting accordingly the bits of the appropriate $d$-degree monomials. In this way, the evolutionary algorithm only explores the restricted search space of homogeneous Boolean functions, which has size $2^{\binom{n}{d}}$. Each individual is evaluated by first retrieving its truth table via the M\"obius transform and then by computing its nonlinearity through the Walsh-Hadamard transform.

\paragraph{Symbolic Encoding.}
The third encoding in our experiments uses tree-based genetic programming (GP) to represent a Boolean function in its symbolic form. In this case, we represent a candidate solution by a tree whose leaves correspond to the input variables $x_1,\ldots, x_n \in \F_2$. The internal nodes are Boolean operators that combine the inputs received from their children and forward their output to the respective parent nodes. The output of the root node is the output value of the Boolean function. The truth table of the function $f: \F_2^n \to \F_2$ is determined by evaluating the tree over all possible $2^n$ assignments of the inputs at the leaves. 

\subsection{Fitness Functions}

\subsubsection{Homogeneous Functions}

The first scenario is concerned only with the difficulty of finding homogeneous functions with the first and third encodings.
To facilitate this, we define a fitness function as a penalty term, which expresses the distance of the evaluated individual from a homogeneous function.
This penalty is defined simply as the number of terms in the ANF form that should not be present in a homogeneous function of the given degree.
Consequently, the fitness function is defined as:
\begin{equation}
\label{eq:fit1}
fit_{1} = -1 \cdot \#\text{\textit{diff\_degree}},
\end{equation}
where $\#$\textit{diff\_degree} represents the number of terms in the ANF of the ``wrong'' degree.
The algorithm assumes maximization, so a fitness value of zero represents a homogeneous function.

\subsubsection{Bent Functions}

Several objective functions can be defined to find bent functions, regardless of the representation and search algorithm. 
The approach used here is based on common choices in related works~\cite{Djurasevic2023} and the authors' previous experience.
Apart from maximizing the nonlinearity value, the applied objective value considers the whole Walsh-Hadamard spectrum and not only its extreme value (see Eq.~\eqref{eq:nonlinearity}); hence, we count the number of occurrences of the maximal absolute value in the spectrum, denoted as $\#max\_values$.
As higher nonlinearity corresponds to a lower maximal absolute value, we aim for as few occurrences of the maximal value as possible, hoping it gets easier for the algorithm to reach the next nonlinearity value.
The algorithm is thus provided with additional information, making the objective space more gradual. The objective value to optimize for bent functions is defined as:
\begin{equation}
\label{eq:bent}
obj_{bent} = nl_{f} + \frac{2^n - \#max\_values}{2^n}.
\end{equation}
The second term never reaches the value of $1$ since, in that case, we effectively reach the next nonlinearity level.

\subsubsection{Homogeneous Bent Functions}

To evolve homogeneous bent functions, we use the objective function value in Eq.~\eqref{eq:bent} and combine it in different ways with the measure of distance from homogeneity.
The first approach in this scenario is first to optimize homogeneity and then add the nonlinearity value only to homogeneous functions.
Following this, the second fitness function is defined as
\begin{equation}
\label{eq:fit2}
fit_2 = fit_1 + \delta_{fit_1, 0} \cdot \left( nl_{f} + \frac{2^n - \#max\_values}{2^n} \right),
\end{equation}
where $\delta_{fit_1, 0}$ is 1 when $fit_1 = 0$ and 0 otherwise. Thus, any positive value denotes a homogeneous function and gives its nonlinearity, while negative values correspond to non-homogeneous functions.

The next approach adds the two terms together, the negative penalty and nonlinearity reward, in the following fitness function:
\begin{equation}
\label{eq:fit3}
fit_3 = fit_1 + \left( nl_{f} + \frac{2^n - \#max\_values}{2^n} \right).
\end{equation}
Thus, we reward the maximization of nonlinearity while allowing the candidate functions outside of the given degree, but all the while decreasing the fitness proportionally.
The value of this fitness function cannot tell us immediately whether an individual represents a homogeneous function. On the other hand, if the fitness value is equal to the nonlinearity of the bent function, then it must also be homogeneous (since attaining greater nonlinearity is not possible).

Finally, the last fitness function to evolve homogeneous bent functions prioritizes nonlinearity and then adds a term to minimize the distance to the homogeneous function of a given degree.
This is defined as follows:
\begin{equation}
\label{eq:fit4}
fit_4 = \left( nl_{f} + \frac{2^n - \#max\_values}{2^n} \right) + \frac{fit_1}{2^n}.
\end{equation}
Since $fit_1$ assumes only negative values (or zero for a homogeneous function), the second term acts as a penalty, which diminishes if the Boolean function assumes the given degree.
Since this penalty is always smaller than 1, the nonlinearity value is considered as a primary criterion.

Due to their definitions, the values of previous fitness functions are not directly comparable. However, if we obtain a fitness value equal to the highest possible nonlinearity, then we know a Boolean function is both bent and homogeneous in a given degree for all three fitness functions.

\subsection{Algorithms and  Parameters}
\label{sec:settings}

\paragraph{Common Experimental Parameters.}
We employ the same evolutionary algorithm for all three encodings: a steady-state selection with a 3-tournament elimination operator (denoted SST). 
In each iteration of the algorithm, three individuals are chosen at random from the population for the tournament, and the worst one in terms of fitness value is eliminated. 
The two remaining individuals in the tournament are used with the crossover operator to generate a new child individual, which then undergoes mutation with individual mutation probability $p_{mut} = 0.5$. The mutated child replaces the eliminated individual in the population.
All experiments use a population size of 500 individuals and the same stopping criterion of $10^6$ evaluations.

\paragraph{Genetic Operators.}
Concerning the genetic operators for both the bitstring and restricted encodings, we use the simple bit mutation and the shuffle mutation. For crossover, we employ one-point and uniform crossover operators. Each time the evolutionary algorithm invokes a crossover or mutation operation, one of the previously described operators is randomly selected.

For the symbolic encoding, we use the following function set: OR, XOR, AND, AND2, XNOR, IF, and function NOT that takes a single argument. This function set is common when dealing with the evolution of Boolean functions with cryptographic properties~\cite{Djurasevic2023}.
The genetic operators used in our experiments with tree-based GP are simple tree crossover, uniform crossover, size fair, one-point, and context preserving crossover~\cite{poli08:fieldguide} (selected at random), and subtree mutation.
The option to use multiple genetic operators was based on previous results indicating better convergence when using a diverse set of operators.

\paragraph{Local Search.}
Finally, all three encodings also employ a generic local search operator, which works in the following way: the operator acts on a single solution and performs a number of mutations. 
If a better solution is found, the new solution immediately replaces the original one, and the operator is applied again.
If no better solution is found after a predefined number of mutations, the operator terminates.
The operator is applied after each generation and acts upon the current best solution and a number of random solutions.
In our experiments, the number of solutions undergoing local search was set to 1\% of the population size, and the number of trials (random mutations per individual) was set to 25.
This operator is general in that it can be applied to any encoding.

\section{Experimental Results}
\label{sec:results}

The experiments are divided into two scenarios: in the first one, we try to obtain homogeneous functions of a given degree. Clearly, in this case, the Genetic Algorithm with restricted encoding (denoted GAr) is not considered since its search space is already constrained to homogeneous functions. Next, in the second scenario, we consider all three encodings to evolve homogeneous bent functions.
The results of the first scenario show that the GA coupled with the bitstring encoding can find homogeneous functions of up to 6 variables only. On the other hand, GP encoding was able to find such a function in at least one run (out of 30) for all function sizes up to 14 variables, where the success rate was 100\% for up to 9 variables.
The results for the second scenario are presented as success rates over 30 runs for each configuration and are shown in Table~\ref{tab:homo_bent}. For the restricted encoding, only the fitness function $fit_3$ has been considered since all other variations have homogeneity penalty terms that are always null in this case. The first finding is that this problem seems very difficult for GA with unrestricted binary encoding since it converges to an optimal solution only a few times for $n=6$ and $d=2$, and not with all fitness functions. Further, considering only quadratic functions, one can see that GAr with the restricted encoding always achieves a full success rate up to $n=14$ variables, while GP manages to find homogeneous bent functions only up to size $n=10$. Finally, a third interesting remark is that there is a steep increase in the difficulty of the problem concerning the degree: already, for $d=3$, no combination of encoding, algorithm, and fitness function has found an optimal solution.

\begin{table}
\small
\caption{Finding $n$-variable homogeneous bent functions of degree $d$ - number of successful runs (out of 30)}
\label{tab:homo_bent}
\centering
\begin{tabular}{@{}lllllllll@{}}
\toprule
                    &        & \multicolumn{3}{c}{GA} & GAr & \multicolumn{3}{c}{GP} \\ \midrule
$n$           & $d$ & $fit_2$   & $fit_3$  & $fit_4$  & $fit_3$ & $fit_2$   & $fit_3$  & $fit_4$  \\ \midrule
\multirow{2}{*}{6}  & 2      & 2      & 8     & 0     & 30      & 25        & 30       & 30    \\
                    & 3      & 0      & 0     & 0     & 0       & 0         & 0        & 0     \\\midrule
\multirow{2}{*}{8}  & 2      & 0      & 0     & 0     & 30      & 30        & 30       & 30    \\
                    & 3      & 0      & 0     & 0     & 0       & 0         & 0        & 0     \\\midrule
\multirow{2}{*}{10} & 2      & 0      & 0     & 0     & 30      & 28        & 30       & 1     \\
                    & 3      & 0      & 0     & 0     & 0       & 0         & 0        & 0     \\\midrule
\multirow{2}{*}{12} & 2      & 0      & 0     & 0     & 30      & 0         & 0        & 0     \\
                    & 3      & 0      & 0     & 0     & 0       & 0         & 0        & 0     \\\midrule
\multirow{2}{*}{14} & 2      & 0      & 0     & 0     & 30      & 0         & 0        & 0     \\
                    & 3      & 0      & 0     & 0     & 0       & 0         & 0        & 0     \\
                    \bottomrule
\end{tabular}
\end{table}

\section{Conclusions and Future Work}
\label{sec:conclusions}

The results reported in this work are still preliminary, but they already provide some insights into the problem of evolving homogeneous bent functions, which could be leveraged for future research on the topic. The most interesting finding is that the problem seems particularly easy to solve by using the unrestricted encoding that forces the solutions to be homogeneous, but only in the quadratic case. A direction worth exploring in this respect is to investigate this phenomenon and verify whether it is due simply to the difference in the size of the search space between different encodings or if more subtle effects related to the shape of the underlying fitness landscapes are at play. Further, it would be interesting to improve the success rates for the degree $3$ (and potentially higher) and understand why the problem is so difficult for any representation except in the quadratic case.

\bibliographystyle{abbrv}
\bibliography{bibliography}

\end{document}